\documentclass[runningheads]{llncs}

\usepackage{accv}

\usepackage{accvabbrv}

\usepackage{graphicx}
\usepackage{booktabs}
\usepackage{multirow}
\usepackage{multicol}
\usepackage[toc,page,header]{appendix}

\newcommand{\bazingadataset}{\textit{Bazinga!}-gold-TV}
\newcommand{\unknown}{\texttt{unknown}}

\usepackage[accsupp]{axessibility}  

\renewcommand \thepart{}
\renewcommand \partname{}

\usepackage{hyperref}

\begin{document}

\title{Character-aware audio-visual subtitling in context} 


\author{Jaesung Huh \and Andrew Zisserman}

\authorrunning{J.Huh and A.Zisserman}

\institute{Visual Geometry Group, Department of Engineering Science, University of Oxford \\
\email{\{jaesung,az\}@robots.ox.ac.uk}}

\maketitle

\begin{abstract} 
This paper presents an improved framework for character-aware
audio-visual subtitling in TV shows.  Our approach integrates speech recognition, speaker
diarisation, and character recognition, utilising both audio and
visual cues.  This holistic solution addresses what is said, when it's
said, and who is speaking, providing a more comprehensive and accurate
character-aware subtitling for TV shows. 
Our approach brings improvements on two fronts: first, we show that
audio-visual synchronisation can be used to pick out the talking face
amongst others present in a video clip, and assign an identity to the
corresponding speech segment. This audio-visual approach improves
recognition accuracy and yield over current methods. Second, we show
that the speaker of short segments can be determined by using the
temporal context of the dialogue within a scene. We propose an
approach using local voice embeddings of the audio, and large language
model reasoning on the text transcription. This overcomes a limitation of
existing methods that they are unable to accurately assign speakers to
short temporal segments.
 We validate the method on a dataset with 12 TV shows, demonstrating superior performance in speaker diarisation and character recognition accuracy compared to existing approaches.
Project page : \url{https://www.robots.ox.ac.uk/~vgg/research/llr-context/}

\keywords{Character-aware audio-visual subtitling \and Audio-visual learning \and Video understanding}
\end{abstract}

\section{Introduction}
\label{sec:intro}

Character-aware audio-visual subtitling is an emerging area that aims
to automatically generate subtitles for TV shows and movies, including
the corresponding speaker names.  This task involves determining three
key aspects: \textit{what} is being said, \textit{when} it is said,
and \textit{who} is saying it. 
This capability is essential for the audio-impaired, so that they can follow video material -- indeed it is a requirement
of Subtitles for Deaf and Hard-of-hearing (SDH~\cite{szarkowska2020subtitling}) that the subtitles include information about
speaker identification, as well as information on sound effects and music. It also enables the
annotation of large-scale video datasets for training 
the next generation of visual-language models,  capable of learning a higher-level story understanding of video material.

The task builds on developments in three specialised areas:
{\em Automatic speech recognition} (ASR, or speech-to-text) 
that is primarily concerned with transcribing spoken words into text -- determining `what is spoken'; {\em Speaker diarisation}, that aims to organise multi-speaker audio into homogeneous single speaker segments, effectively solving `who spoke when'; and {\em Character recognition}, that aims to identify the characters appearing in the video clips.
Each of these areas is well explored, and can use single modality methods (\ie audio only or visual only) or audio-visual methods. 
For example, ASR can be audio-only~\cite{radford2023whisper, hsu2021hubert,
gulati2020conformer}, or audio-visual~\cite{Afouras19, shi2022avhubert,
ma2023auto, gabeur2022avatar}. 
Similarly, common methods can be used across the areas. 
For example, voice embeddings can be used for diarisation by clustering~\cite{wang2018speaker, zhang2019fully, kinoshita2021integrating}, and for recognition by matching to a gallery of voices~\cite{li1982text, suchitha2015feature, kaphungkui2019text}. 
However, because these are somewhat independent areas, they do not alone provide all the ingredients required.

Recent works have introduced methods and datasets for character-aware
audio-visual subtitling, building on elements from the three areas
above~\cite{Korbar24,lerner2022bazinga}. 
The state-of-the-art method of Korbar {\it et al.}~\cite{Korbar24}, proceeds in two stages: it first builds a gallery of voice embeddings for each character using audio-visual methods, and then generates the character-aware subtitles using only audio recognition. 
Despite its accomplishments, however, this method has two significant shortcomings when determining the character speaking during a temporal segment: (1) it has a poor performance for short segments
(those lasting less than 2 seconds), often assigning the wrong character; and (2) it has a low yield over all segments, as often it is unable to classify the character.

\begin{figure}[t]
  \centering
  \includegraphics[width=\linewidth]{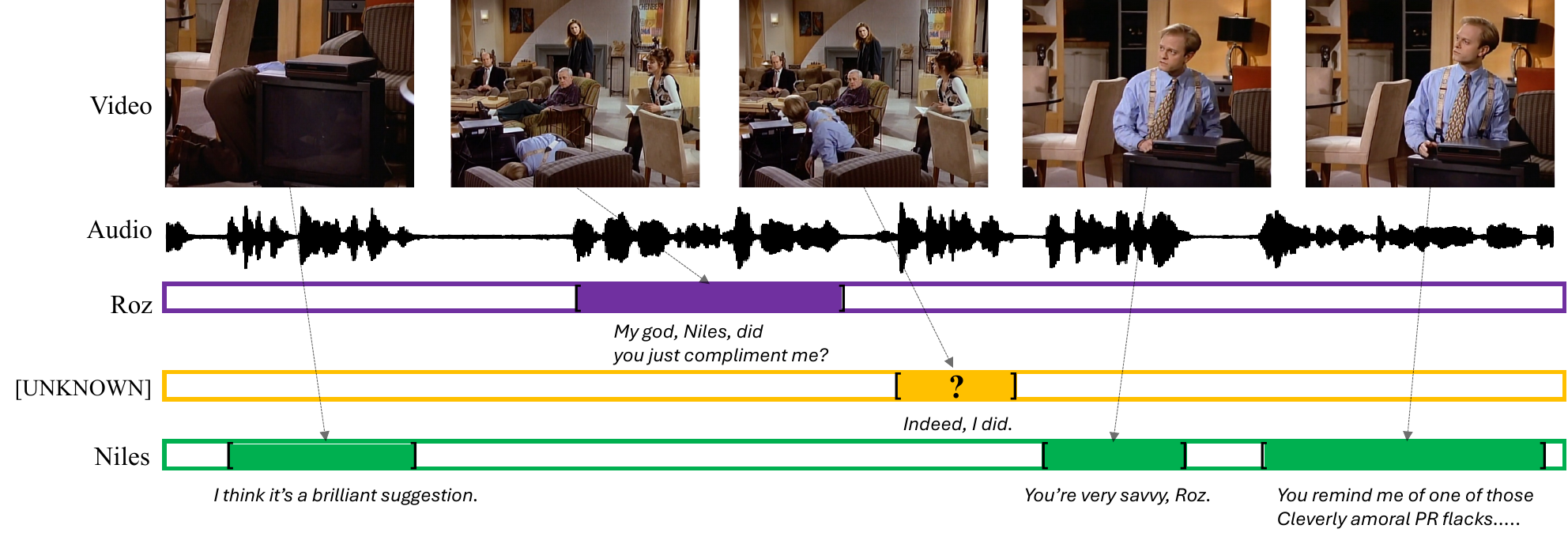}
  \caption{An example video clip and output of our method. 
Dialogues in TV shows typically flow continuously, and speaker identities can often be
inferred from the content and context of the conversation.  
In some cases, it's possible to diarise speakers solely based on textual context. 
Even though we cannot see the speaker visually -- so have no evidence from lip-movement -- we can infer that the utterance with a question mark (\textbf{?}) belongs to `Niles' by looking at temporal context of the dialogue.}
  \label{fig:temporal_context}
\end{figure}

In this paper we make three contributions. 
First, we introduce a new method for identifying the speaker for short segments, building on the insight that assignments that cannot be resolved using only local (temporal) information, can often be disambiguated using the {\em temporal context}  of the surrounding dialogue. 
We investigate two complementary approaches for this task: (i) using speaker recognition, 
we note that a short utterance in a dialogue may well be spoken by a character with a longer utterance  (where the identity is not ambiguous) elsewhere in the scene, and the short segment can then be assigned using {\em local} voice embeddings, rather than voice embeddings from a gallery where audio conditions may substantially differ; (ii) using a large language model (LLM), the identity of the character speaking the short segment can be resolved based on the {\em content} of the dialogue, as illustrated by the example in Figure~\ref{fig:temporal_context}.
The second contribution is to use a {\em local visual embedding} around the lip motion synchronised with the speech to determine the speaker. 
This overcomes a limitation of~\cite{Korbar24}, where a CLIP descriptor of the entire frame is used to predict the speaker identity. 
The use of a local visual embedding leads to higher yield of assignments for speaker segments. 
Taken together these two contributions significantly improve the performance over that of~\cite{Korbar24}. 
As our third contribution, we validate our method on a large evaluation dataset covering 12 TV series. 
This dataset incorporates the existing dataset used by~\cite{Korbar24} and additional shows from~\cite{lerner2022bazinga}, demonstrating the generalisation ability of our method. 
\section{Related Work}
\label{sec:relatedworks}

Several subtasks within this field have already been explored by researchers. 
\textit{Speech recognition}~\cite{Chan15, gulati2020conformer, malik2021automatic, radford2023whisper}, or speech-to-text, is primarily concerned with transcribing spoken words into text. 
However, this subtask typically overlooks the timing of speech and fails to identify the speaker.
\textit{Speaker diarisation}~\cite{wang2018speaker, diez2019analysis, Chung20,park2022review} aims to identify speech regions and assign speaker labels to each person in an audio file. 
This task clusters speech segments by speaker without necessarily matching them to specific known individuals.
\textit{Character recognition}~\cite{Berg04a, ramanathan2014linking, Kalogeiton20, hu2015deep, poignant2017multimodal}, a well-studied topic in computer vision and speech processing, assigns names to characters appearing in scenes.
Character-aware audio-visual subtitling requires the integration of all three tasks, utilising both audio and visual cues from the video. 

\paragraph{Character recognition in videos.}
Recognising characters in video~\cite{Everingham06a, Everingham09, Nagrani17b, haurilet2016naming} is a challenging task due to the presence of multiple characters in a single frame, occlusions, and variations in appearance.
Several methods have been proposed to incorporate additional modalities, such as audio~\cite{Brown21, Nagrani17b}, or transcripts~\cite{Everingham06a, Everingham09, Bojanowski13, haurilet2016naming} which are often unavailable.
There are a line of works which use speaker diarisation in TV shows and use the result to cluster the speaker identities~\cite{bost2015audiovisual, sharma2022using}. 
However, they simply cluster the speaker identities, not assigning the actual character's name.
Our task involves assigning the specific names of speakers in TV shows using a castlist.

\paragraph{Audio-visual speech processing.}
Numerous studies have examined human conversation from a broad perspective. 
Given that these interactions primarily occur through speech, a wide range of research focuses on audio-only approaches to various tasks, including speech recognition~\cite{radford2023whisper, baevski2020wav2vec, gulati2020conformer}, speaker identification~\cite{desplanques2020ecapa, chung2020in, koluguri2022titanet}, and speaker diarisation~\cite{bredin2023pyannote, Fujita2019Interspeech}.
With a rise of the multimodal learning, researchers have started to incorporate visual information such as lip movements in addition to audio information to improve the performance of these tasks.
For example, they use lip movements~\cite{Afouras19, shi2022avhubert} or faces~\cite{Chung20,Brown21c} in addition to audio to improve the performance of this task.

\paragraph{LLM for video understanding.}
Large Language Models (LLMs)~\cite{llama3, achiam2023gpt4, jiang2023mistral, team2023gemini} have driven the great progress not only in Natural Language processing but also in computer vision~\cite{Qwen-VL, liu2023llava, alayrac2022flamingo, lu2024deepseekvl} and audio processing~\cite{gong_ltuas, gong2023listen}.
Over the past few years, there has been a plethora of works which leverage LLMs in various video understanding tasks.
There are two different approaches to this. 
The first approach integrates a pretrained LLM with visual or audio backbones as part of the entire model, fine-tuning it to understand multimodal content~\cite{damonlpsg2023videollama, song2023moviechat, Han23, Maaz2023VideoChatGPT}.
The second approach uses LLM separately from video models to improve performance on video understanding tasks~\cite{chen2023video, wang2023chatvideo}.

\paragraph{Human conversation datasets.}
There has been growing interest in audio-visual datasets with rich transcriptions of spoken conversations, including speech transcripts, timestamps and speaker identities. 
Several datasets exist with annotations for either one of these aspects.
LRS series~\cite{Afouras19, afouras2018lrs3} have advanced audio-visual speech recognition technology, but their single-speaker focus limits development of multi-speaker systems for conversational settings.
There exist audio-visual speaker diarisation datasets~\cite{Chung20,xu2022avaavd} with multiple speakers but do not have a speech transcripts.
The AMI-Corpus~\cite{kraaij2005ami} and VoxMM~\cite{kwak2024voxmm} are multimodal datasets which provide audio-visual data with speaker identities and speech transcripts.
However, both focuses on different domain than ours such as meeting scenarios, commercial or interviews. 
\textit{Bazinga!}~\cite{lerner2022bazinga} offers rich transcriptions of TV shows, including word-level timestamps, speech transcripts, and speaker identities. We use this dataset to verify our pipeline's performance.

\paragraph{Relation to the method of Korbar {\it et al.}\cite{Korbar24}.} 
This work also aims to generate character-aware subtitle generation.
However, it has several limitations. 
Firstly, it fails to utilise spatial information from lip-moving areas, which could significantly enhance speaker recognition accuracy. 
The method utilises CLIP-PAD~\cite{Korbar22} which recognises characters in scenes without employing a face detection model to identify clips for single-speaker regions. 
Unfortunately, these clips may contain multiple faces, potentially confusing the model when tasked with identifying the actual speaker.
Secondly, it doesn't take advantage of the time-based context when matching speaker names to parts of speech. 
In TV shows, conversations usually progress in a continuous manner. 
As a result, it's often possible to figure out who is speaking by considering the overall flow of the dialogue and the context in which things are said.

\section{Assigning Speakers to Short Audio Segments}
\label{sec:method-audio}

The task here is to assign speaker identities to short temporal speech segments. We assume that we have a gallery/library of voice embeddings available for the principal characters.

\begin{figure}[t]
  \centering
  \includegraphics[width=\linewidth]{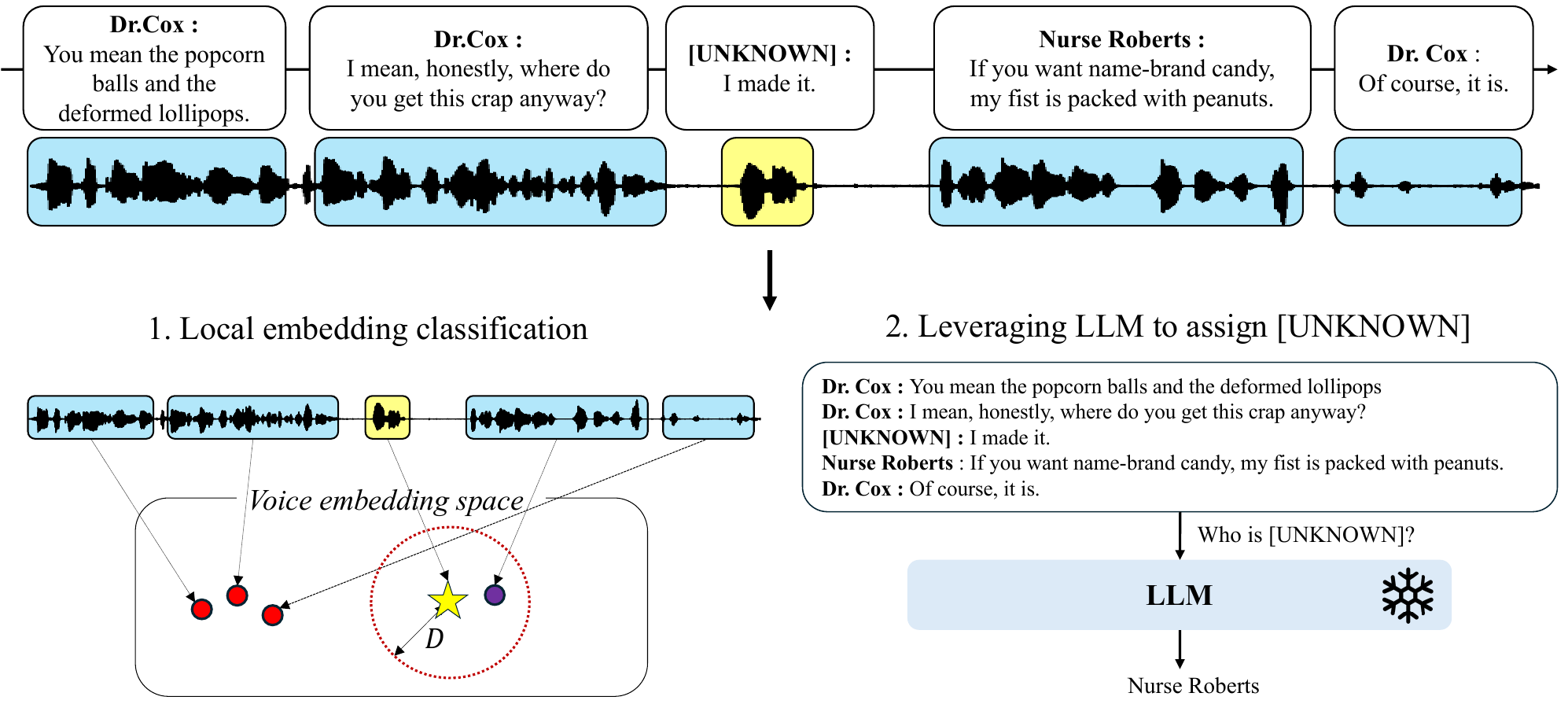}
  \caption{{\bf Assigning speakers to short audio segments}. First, we use speaker embeddings from nearby segments where we have high confidence in speaker identification (left). Second, we employ a Large Language Model (LLM) to determine the speaker based on the content of conversation. (right)}
  \label{fig:short}
\end{figure}

It is well known that identifying speakers by their voice alone typically fails in verification tasks when the input audios are short~\cite{poddar2018speaker, he2023voiceextender}. This is because
state-of-the-art speaker embedding extractors~\cite{desplanques2020ecapa, koluguri2022titanet, torgashov2023id} are normally trained with segments of at least 2 seconds of audio waveforms.
TV shows contain many short segments (see \cref{fig:length_distribution}), resulting in false classification when using standard classification methods on the embedding, such as nearest class centroid.

To solve this short segment speaker assignment problem, we use the {\em temporal context} of the human conversation.
There is a high chance that the speaker of the short speech segment we are interested in is involved in the dialogue : which means that the speaker might speak elsewhere and for longer within the scene.
The key concept is that speaker identity can be accurately predicted for longer audio segments. These identified segments can then be used to classify speakers in shorter audio segments nearby.
We employ this idea into two complementary ways: using local speech embeddings, and using the language (text) of the dialogue.
Assuming we know the speakers of long audio segments with a high confidence, we demonstrate how to leverage this information to determine the speakers in shorter audio segments.
The method of using temporal context is illustrated in \cref{fig:short}.
\paragraph{Local embedding classification.}
\label{subsec:method_local}

As is known from diarisation, there are advantages in comparing to {\em local} embeddings when deciding if two speakers are the same or not. Since the two embeddings are computed under the same environment -- and so have the same background sounds, the same reverberation, even the same microphone, many of the `nuissance' variables are removed, simplifying the classification challenge. Thus, to determine the speaker of a short segment, its embedding is compared to the segment embeddings of the other (known) speakers in the scene, instead of comparing to their class centroid. 

In detail, we extract the speaker embeddings within $n_{local}$ preceding and succeeding sentences around the segment of interest (where $n_{local} = 15$).
Then the speaker is assigned by computing the distance between the embeddings of the short segment and segments with known speakers using first nearest neighbor classification. If the distance is below a threshold $D$ then the assignment is accepted, otherwise the short segment is classed as \unknown, and the assignment is determined (if possible) by using the text content, as described next.

\paragraph{Leverage LLM to assign speakers of} \unknown.
\label{subsec:method_llm}
As illustrated in \cref{fig:temporal_context},  the speaker identity can be inferred solely by using the {\em content} of the dialogue (\ie without actually hearing the voice). 
Since large language models (LLMs) have a good predictive `understanding' of dialogues, they can be queried to predict the speaker of the short segment, given the named speakers of other utterances in the dialogue. 
We apply this LLM classification in the cases that cannot be classified using voice alone, since it is a weaker cue. 

Specifically, we ask the LLM model to predict who the speaker is of the short segment, using zero-shot prompting. 
We provide the $n_{llm}$ (\eg 15) sentences with the speaker names both before and after this \unknown\ sentence.
The LLM model is tasked to answer with: either one of the characters that appear within this dialogue with $2n_{llm}+1$ sentences; or \unknown\ if the speaker is from outside the dialogue or if the speaker cannot be inferred only from the provided dialogue.
The prompt used also has three examples along with their answers, followed by the actual query and dialogue.
The detailed prompt instruction is provided in the supplementary material.

\section{Using Local Visual Predictions to Assign Speakers}
\label{sec:method-visual}
\begin{figure}[t]
  \centering
  \includegraphics[width=\linewidth]{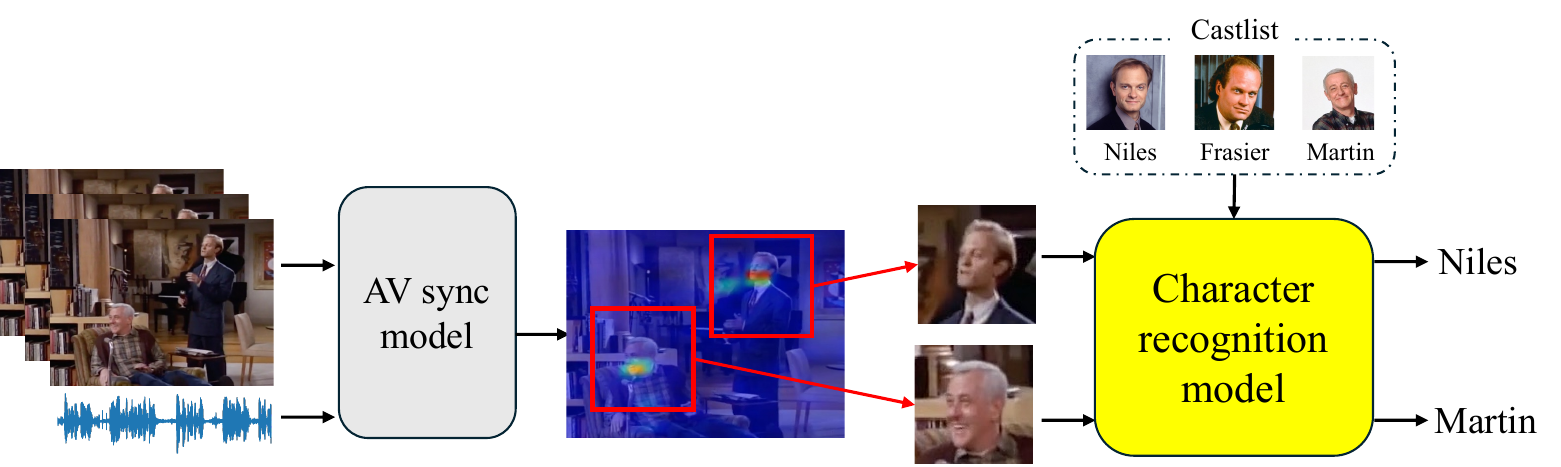}
  \caption{The visual prediction process for a speech segment. Visible speakers with lip movements synchronised with the speech audio are recognised by using a visual embedding from the castlist. This assigns an identity to the corresponding speaker.}
  \label{fig:vis_pred}
\end{figure}

The task here is to recognise all characters speaking within a video clip.
We assume that we have a gallery/library of visual embeddings available for the principal characters. 
Although speech is sometimes difficult to recognise due to background noise or overlapping voices, the corresponding visual frames often provide a clear view of the speaker. 
We can use this visual information to help identify speakers.  \cref{fig:vis_pred} illustrates the method.

To identify all visible speakers in the scene, we employ a multi-step process. 
First, we run a pretrained audio-visual synchronisation model~\cite{Afouras20c} that detects lip motions by producing a heat map where areas around moving lips are activated. 
We then crop spatial regions around the detected peaks with a fixed width and height, and extract visual embeddings from each cropped region using a CLIP-based character recognition model~\cite{Korbar22}. 
We compare the distances between these visual embeddings to all actors in the cast list. 
Finally, we select the cropped regions that identify speakers with high confidence (above a predetermined threshold) and store these predictions for subsequent speaker assignment steps.

This cropping approach essentially extracts a local visual embedding of the face. It overcomes a limitation of the  character recognition model~\cite{Korbar22}, which is confused if multiple people appear in the same frame since it uses a global frame embedding.
In summary, we use audio-visual cues to assign identities to speaking segments with visible faces, and audio-only embeddings to assign identities when the face is not visible.

\section{Implementation Details}
\label{sec:method-details}

\begin{figure}[t]
  \centering
  \includegraphics[width=\linewidth]{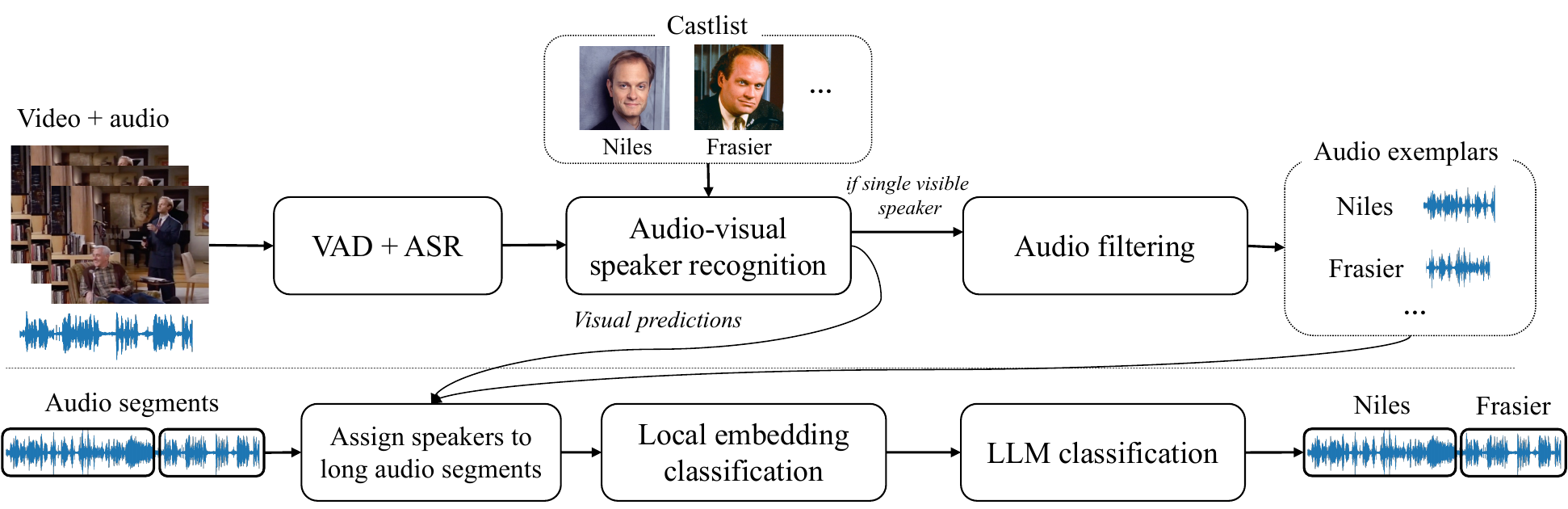}
  \caption{A schematic overview of our pipeline. We first extract the audio exemplars from videos (top) and use them label all audio segments (bottom).}
  \label{fig:overview}
\end{figure}

We follow the approach of~\cite{Korbar24} for generating character-aware subtitles using video and a cast list for each show. 
Their two-stage method first builds a gallery of audio exemplars -- speech segments with high-confidence character name assignments. 
These exemplars are then used to assign speaker names to all speech segments using centroid classification. 
If the minimum distance from the nearest exemplar exceeds a threshold, then no specific character name is assigned, allowing for characters without exemplars who cannot be classified.
We detail our implementation of this method in the following subsections, highlighting the improvements we have made over the original method of~\cite{Korbar24}.
\cref{fig:overview} shows a schematic overview of our entire pipeline.

\subsection{Stage 1. Building audio exemplars}
\label{subsec:method_stage1}
The goal of this stage is to extract audio exemplars from the video for which we know the corresponding speaker.

We first run Voice Activity Detection (VAD) based on Automatic Speech Recognition (ASR) model on the audios to generate the speech transcripts with corresponding timestamps at a sentence level.
Visible speakers are then determined by using the synchronisation of lip movements and the speech with the self-supervised trained audio-visual model~\cite{Afouras20b} that produces a heatmap of where the lip-motions are synchronised.
We crop the surrounding spatial regions of each peaks in the heatmap and visually recognise characters in the region.
Video clips with a single peak are kept as exemplar candidates, but predictions from the clips with more than one peak are also kept for assigning the speaker later.
Then, we conduct additional audio filtering for the exemplar candidates to reduce the label noise.
We detail the process below.

\subsubsection{Stage 1--1. VAD + ASR.}
\label{subsubsec:method_exemplar_vadasr}
The goal of this stage is to generate speech transcripts with corresponding timestamps.
We use publicly available pipeline~\cite{Bain23} to produce speech transcripts with timestamps in a sentence level. We assume each sentence is spoken by a single speaker at this point, but we address the case of overlapping speech in subsequent stages. This step produces subtitles without speaker identities.
Unlike \cite{Korbar24}, we do not use any pretrained laughter detector. 
Instead, we run a speech enhancement network to reduce background noise in the following step. 

\subsubsection{Stage 1--2. Audio-visual speaker recognition.}
\label{subsubsec:method_exemplar_av}
This stage aims to recognise all speakers in the visual scenes and collect video clips with a single visible speaker using a castlist per video.
First, we run a pretrained audio-only speech enhancement model~\cite{defossez2020denoiser} to reduce background noise, thereby reducing false positives in the following stage. 
Then, we visually recognise the visible speakers' identities by using the method explained in \cref{sec:method-visual}.
We need a gallery of images to compare the distance between each visible person and characters in the castlist.
We collect up to 10 images per each character and form a visual embedding per character.

After running this model, we categorise the video clips into three types: (i) clips without any peaks, (ii) clips with a single peak, and (iii) clips with multiple peaks. 
The second type, clips with a single peak, are considered as our exemplar candidates in subsequent stages.
However, we proceed to run the character recognition model on both the second and third types.
We keep the output predictions to use as candidates for classification in Stage 2.
Multiple peaks can occur for two reasons: either multiple speakers are talking simultaneously, or the model produces false positives.

\subsubsection{Stage 1--3. Audio filtering}
\label{subsubsec:method_exemplar_audiofilter}
This stage further reduces label noise by focusing on single-speaker video clips from the previous stage. 
We extract speaker embeddings from the audio and analyse each embedding's $N$ (\eg 5) nearest neighbors. 
We retain the embedding only if all $N$ neighbors belong to the same speaker; otherwise, we remove the corresponding audio segment from our exemplars. 

\subsection{Stage 2. Assigning speaker identities of each speech segment}
\label{subsec:method_stage2}
Stage 1 aims to collect audio exemplars for which we know the corresponding speaker identities with high certainty. 
This stage aims to assign speaker identities to all segments.
We first classify the long audio segments ($>$ 2 sec) and segments with extreme high confidence.
For each segments, we only compare the distance between the exemplars from the visible speakers, which we obtain in Stage 1--2. 
If no visible speakers are detected, we compare the distance between the exemplars from castlist.
Then, we use the local temporal context, using local embedding classification and LLM which are described in \cref{sec:method-audio}.
The long segments as well as audio exemplars from the previous stage are used for local embedding classification.

After assigning speakers to each audio segments in a sentence level, we run the public overlapping speech detection model to detect the overlapping speech.
If the segment is detected with overlapping speech, we assign the speaker with two nearest speakers along the time axis.

\subsection{Implementation details}
\label{subsec:method_detail}
We use WhisperX~\cite{Bain23} for VAD+ASR model. 
We further use Silero VAD~\cite{Silero_VAD} for filtering out the false detections from the WhisperX.
ECAPA-TDNN~\cite{desplanques2020ecapa} is used for speaker embedding extractor, pretrained with VoxCeleb~\cite{Nagrani19}.
We use LWTNet~\cite{Afouras20b} for audio-visual synchronisation model and crop the activate region with $W=H=350px$ to recognise the characters.
We use $n_{local} = 15$ preceding and suceeding audio segments.
We use public official Llama3-70B 4-bit quantised model, finetuned with instruction sets, to assign speakers with $n_{LLM}=15$ preceding and succeeding sentences.
We use public overlapping detection model from \texttt{pyannote 2.1}~\cite{Bredin2021overlap}.
Rest of the parameters are identical to those in \cite{Korbar24}. 
After detecting overlapping speech, we divide the audio segments wherever there is silence longer than 1 second, using word-level timestamps from WhisperX.
The same speaker is assigned to these divided audio segments.
All hyperparameters are determined by grid search on validation sets.

\section{Dataset and evaluation metrics}
\label{sec:exp}
This section explains the dataset we have used to validate our method and evaluation metrics. 

\subsection{Dataset}
\label{subsec:exp_dataset}
\subsubsection{LLR-TV:}
\cite{Korbar24} has released a TV shows dataset including six episodes each from Frasier, Seinfeld, and Scrubs, along with transcripts, speaker names, and timestamps. 
The official dataset website~\footnote{\url{https://www.robots.ox.ac.uk/~vgg/research/look-listen-recognise/}} has released version 1.1, which includes fixed annotations they have made. 
We use the current version to verify our method, but we also report the performance from the original paper in \cref{sec:results}. 
For each series, we use the sixth episode as our validation set to determine the hyperparameters. 
The rest of the dataset is used as our test set.

\subsubsection{\textit{Bazinga!}-gold-TV:}
\label{subsubsec:exp_dataset_llr}

\textit{Bazinga!}~\cite{lerner2022bazinga} dataset provides a rich set of annotations from 16 different TV shows and movies, such as speech transcripts with timestamps, speaker, addressee and entity linking information.
The dataset itself is divided into gold and silver based on the level of annotations.
We use all TV shows in the gold set to verify our method including \textit{Battlestar Galactica (B.G.)}, \textit{Breaking Bad (B.B.)}, \textit{Buffy the Vampire Slayer (Buffy)}, \textit{Friends}, \textit{Game of Thrones (GoT)}, \textit{Lost}, \textit{The Big Bang Theory (TBBT)}, \textit{The Office (Office)} and \textit{The Walking Dead (W.D.)}.
We exclude \textit{StarWars} since our paper focuses on TV shows.

Since our method is audio-visual while the dataset only provides audio, we need to adjust the timestamps in the annotations to match our videos. 
We use the audio-audio alignment method introduced by \cite{Han24} to obtain precise temporal alignment by comparing the audio provided in the dataset with the audio from our video source. 
Similar to LLR-TV, we use the last episode of each series as our validation set, with the remaining episodes serving as our test set.

\begin{figure}[t]
  \centering
  \includegraphics[width=0.85\linewidth]{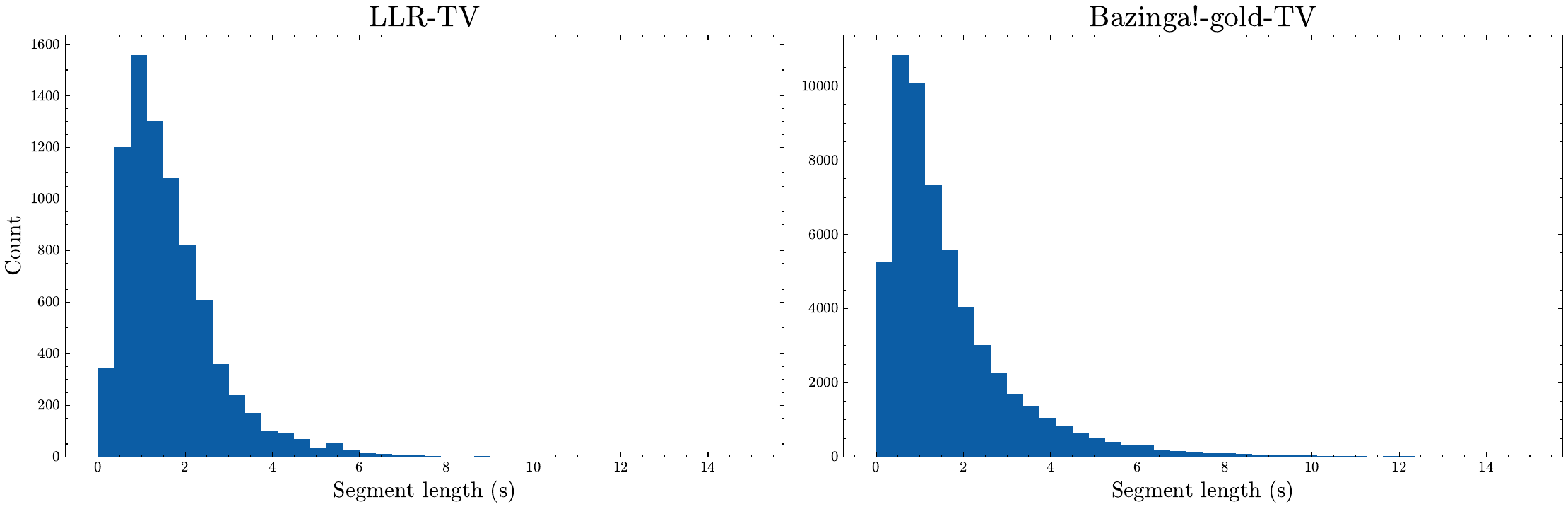}
  \caption{Distribution of segment lengths on LLR-TV and \bazingadataset.}
  \label{fig:length_distribution}
\end{figure}

\cref{fig:length_distribution} shows the distribution of segment lengths in both datasets. 
71.0\% of segments in LLR-TV and 71.5\% in \bazingadataset\ are shorter than 2 seconds. 
This indicates that recognising speakers in shorter segments is crucial for analysing conversations in TV shows.
Note that while LLR-TV is manually corrected, \bazingadataset\ provides timestamps obatined through force-alignment.
Thus the annotation is relatively noisy (\eg they do not provide annotations for \textit{Previously...} part.).

\subsection{Evaluation metrics}
\label{subsec:exp_metrics}

\subsubsection{Diarisation metrics.}
\label{subsusbsec:exp_metrics_diar}
\textbf{DER}~\cite{nistrt09} is a standard evaluation metric for speaker diarisation. 
However, recent studies~\cite{cheng2022cder, liu2022ber} have highlighted a significant limitation of DER: its time-duration-based computation fails to accurately capture the recognition performance for short-term segments. 
To address this limitation, Conversational-DER (\textbf{CDER}) is introduced, which calculates speaker diarization accuracy at the utterance level and also accounts for short segments. 
For more details on CDER, please refer to \cite{cheng2022cder}.

In this paper, we employ DER with a forgiveness collar of 0.25 seconds, taking into consideration instances of overlapping speech. 

\subsubsection{Character recognition metrics.}
\label{subsusbsec:exp_metrics_char}
Character recognition accuracy (\textbf{Acc.}) is calculated for segments that overlap with ground truth segments.
A segment is considered correctly classified if the character's name is accurately identified and matches the corresponding overlapping ground truth segment.
Precision and recall for character identification are also reported for each show.
Both metrics are calculated for all characters (\textbf{Prec.} and \textbf{Rec.}) and separately for main characters (\textbf{Prec.(M)} and \textbf{Rec.(M)}) in each series.
A list of main characters for each show is provided in the supplementary material.

\section{Results}
\label{sec:results}

This section presents our overall results on the test sets, comparing them to other baselines. 
We also provide a detailed analysis of how our method accurately collect audio exemplar. 
We conclude by showcasing qualitative examples of our method and comparing them to the baseline.
The effect of using local visual predictions and speech enhancement is shown in the supplementary.

\subsection{Overall performance}
\label{subsec:results_overall}
\paragraph{Diarisation performance.}

\begin{table}[t]
\centering
\caption{Diarisation performance on LLR-TV test set. Lower is better. \textbf{LLR}$*$ is from the original paper before the GT was corrected and \textbf{LLR}$\dagger$ is our reproduced result with annotation corrections from the website. \textbf{DER} : Diarisation Error Rate (\%), \textbf{CDER} : Communication DER (\%), \textbf{A} : Audio, \textbf{V} : Video.}
\label{tab:diarization_comparison}
\resizebox{0.8\linewidth}{!}{
\begin{tabular}{@{}lccccccc@{}}
\toprule
\multicolumn{1}{c}{}      & \textbf{}         & \multicolumn{2}{c}{\textbf{Frasier}} & \multicolumn{2}{c}{\textbf{Seinfeld}} & \multicolumn{2}{c}{\textbf{Scrubs}} \\ \midrule
\multicolumn{1}{c}{Model} & \textbf{Modality} & \textbf{DER}      & \textbf{CDER}    & \textbf{DER}      & \textbf{CDER}     & \textbf{DER}     & \textbf{CDER}    \\ \midrule
SimpleDiar~\cite{simplediarization}         & \textbf{A}        & 24.2              & 58.5             & 24.5              & 56.2              & \textbf{24.4}    & 52.6             \\
\texttt{pyannote}~\cite{bredin2023pyannote}                  & \textbf{A}        & 24.7              & 84.1             & 35.4              & 88.7              & 31.1             & 75.8             \\
LLR$*$~\cite{Korbar24}                      & \textbf{A + V}    & 24.3              & -                & 29.7              & -                 & 36.4             & -                \\
LLR$\dagger$~\cite{Korbar24}                       & \textbf{A + V}    & 26.4              & 39.1             & 28.0              & 40.7              & 26.7             & 40.3             \\
Ours                      & \textbf{A + V}    & \textbf{20.3}     & \textbf{28.8}    & \textbf{23.3}     & \textbf{33.6}     & 25.7             & \textbf{37.0}    \\ \bottomrule
\end{tabular}}
\end{table}

\begin{table}[t]
\centering
\caption{Diarisation performance on \bazingadataset. \textbf{DER} : Diarisation Error Rate (\%), \textbf{CDER} : Communication DER (\%), \textbf{Mod} : Modality, \textbf{A} : audio, \textbf{V} : Video.}
\label{tab:diar_bazinga}
\resizebox{\linewidth}{!}{
\begin{tabular}{@{}lccccccccccccccccccc@{}}
\toprule
\multicolumn{1}{c}{}               & \textbf{}    & \multicolumn{2}{c}{\textbf{B.G.}} & \multicolumn{2}{c}{\textbf{B.B.}} & \multicolumn{2}{c}{\textbf{Buffy}} & \multicolumn{2}{c}{\textbf{Friends}} & \multicolumn{2}{c}{\textbf{GoT}} & \multicolumn{2}{c}{\textbf{Lost}} & \multicolumn{2}{c}{\textbf{TBBT}} & \multicolumn{2}{c}{\textbf{Office}} & \multicolumn{2}{c}{\textbf{W.D.}} \\ \midrule
\multicolumn{1}{c}{\textbf{Model}} & \textbf{Mod} & \textbf{DER}    & \textbf{CDER}   & \textbf{DER}    & \textbf{CDER}   & \textbf{DER}     & \textbf{CDER}   & \textbf{DER}      & \textbf{CDER}    & \textbf{DER}    & \textbf{CDER}  & \textbf{DER}    & \textbf{CDER}   & \textbf{DER}    & \textbf{CDER}   & \textbf{DER}     & \textbf{CDER}    & \textbf{DER}    & \textbf{CDER}   \\ \midrule
SimpleDiar~\cite{simplediarization}                         & \textbf{A}   & 61.3   & 101.0           & 68.8            & 154.5           & 31.5             & 62.6            & \textbf{46.8}              & 85.6             & 38.3            & 85.8           & 90.9            & 117.2           & \textbf{20.6}            & 38.0   & \textbf{33.3}    & \textbf{70.3}    & 93.4            & 123.9           \\
\texttt{pyannote}~\cite{bredin2023pyannote}                           & \textbf{A}   & \textbf{58.7}            & 104.9           & 70.4            & 136.7           & \textbf{31.3}    & \textbf{59.6}   & 60.5              & 128.8            & \textbf{37.2}   & 85.1           & \textbf{88.1}   & 111.9           & 30.3            & 70.0            & 35.4             & 112.5            & 100.7           & 138.2           \\
LLR$\dagger$~\cite{Korbar24}                                & \textbf{A+V} & 79.5            & 101.6           & 92.9            & 135.4           & 55.4             & 77.5            & 55.9              & 72.6             & 63.7            & 120.0          & 111.7           & 115.4           & 29.5            & 39.5            & 44.2             & 93.9             & 108.9           & 130.6           \\
\textbf{Ours}                      & \textbf{A+V} & 62.7   & \textbf{87.2}   & \textbf{67.0}   & \textbf{99.4}   & 46.2             & 62.0            & 47.1     & \textbf{64.4}    & 44.2            & \textbf{82.7}  & 89.0            & \textbf{86.6}   & 27.3   & \textbf{36.0}   & 40.2             & 71.8             & \textbf{92.5}   & \textbf{97.1}   \\ \bottomrule
\end{tabular}}
\end{table}

We report the diarisation performance of our method on the LLR-TV test set in \cref{tab:diarization_comparison}. 
Our method is compared against three competitive baselines, including two audio-only models and one audio-visual diarisation method, LLR.
In terms of DER, our method demonstrates superior performance on Frasier and Seinfeld compared to all other models, and achieves comparable results on Scrubs. 
More notably, when considering the CDER, our method significantly outperforms other baselines with margins of 10.3\%, 7.1\%, and 3.3\% on Frasier, Seinfeld, and Scrubs, respectively. 
This indicates that our method recognises characters more accurately, even in short segments, compared to other methods.

\cref{tab:diar_bazinga} compares our pipeline against other baselines on the \bazingadataset, where our method shows better performance in most TV series. 
The metrics should be taken with a `grain of salt', a point that is also made in the original paper~\cite{lerner2022bazinga}.

\paragraph{Character recognition performance.}

\begin{table}[t]
\centering
\caption{Character recognition performance on LLR-TV test set. \textbf{Prec.} and \textbf{Rec.} indicate the precision and recall of overall audio segments respectively, while \textbf{Prec.(M)} and \textbf{Rec.(M)} are those of main characters in TV shows. \textbf{Acc.} is the character recognition accuracy for those which overlap with one of the groundtruth timestamps.}
\label{tab:characc_llr}
\resizebox{\linewidth}{!}{
\begin{tabular}{@{}lccccc|ccccc@{}}
\toprule
                  & \multicolumn{5}{c|}{LLR~\cite{Korbar24}}                                                                                               & \multicolumn{5}{c}{\textbf{Ours}}                                                                                             \\ \midrule
                  & \textbf{Acc.} & \textbf{Prec.} & \textbf{Rec.} & \multicolumn{1}{l}{\textbf{Prec.(M)}} & \multicolumn{1}{l|}{\textbf{Rec.(M)}} & \textbf{Acc.} & \textbf{Prec.} & \textbf{Rec.} & \multicolumn{1}{l}{\textbf{Prec.(M)}} & \multicolumn{1}{l}{\textbf{Rec.(M)}} \\ \midrule
\textbf{Frasier}  & 87.0           & \textbf{91.6}  & 87.0          & \textbf{92.5}                         & 89.4                                  & \textbf{88.9} & 89.8           & \textbf{88.9} & 90.3                                  & \textbf{92.6}                        \\
\textbf{Seinfeld} & 84.5           & \textbf{89.0}  & 84.6          & \textbf{92.5}                         & 89.4                                  & \textbf{85.8} & 87.1           & \textbf{86.0} & 89.5                                  & \textbf{90.7}                        \\
\textbf{Scrubs}   & 84.3           & \textbf{89.2}  & 84.9          & \textbf{91.0}                         & 88.1                                  & \textbf{84.4} & 84.8           & \textbf{85.1} & 85.1                                  & \textbf{90.7}                        \\ \bottomrule
\end{tabular}}
\end{table}

\begin{figure}[t]
  \centering
  \includegraphics[width=0.95\linewidth]{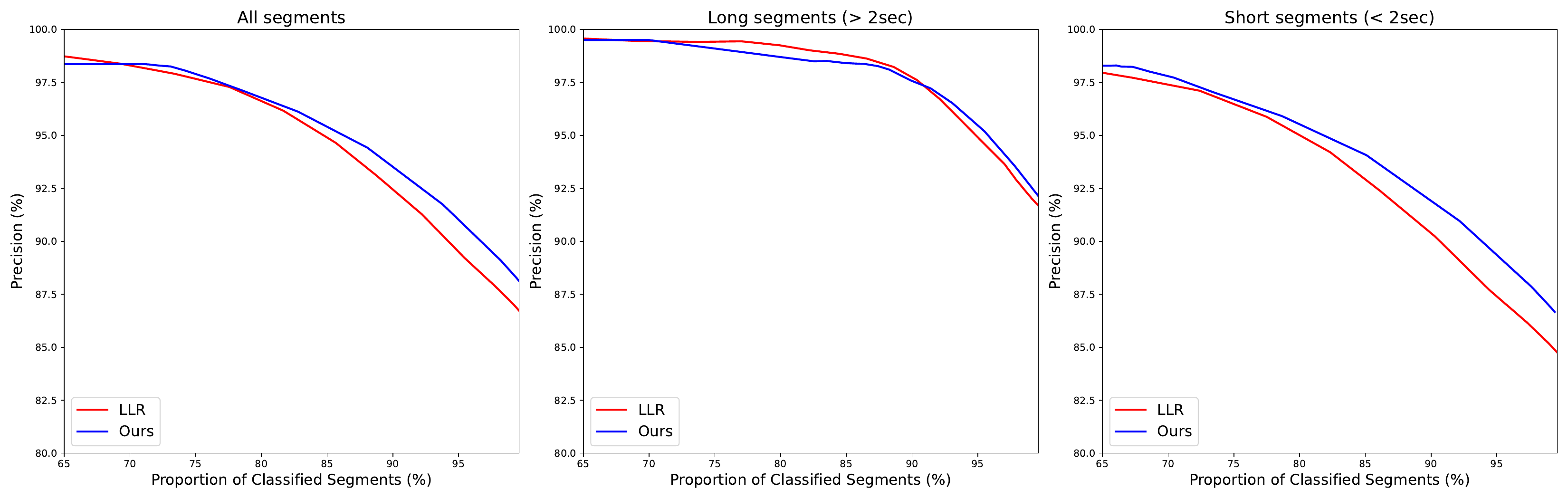}
  \caption{Precision-POCS curves for audio segments in LLR-TV test set.}
  \label{fig:precision_pocs}
\end{figure}

\cref{tab:characc_llr} shows the character recognition performance on LLR-TV.
The reported performance from both LLR and our method is based on the highest accuracy achieved by varying the hyperparameter $D$ (see \cref{subsec:method_local}) on the validation set.
Compared to the reproduced LLR, our method demonstrates higher character recognition accuracy.
Interestingly, although the LLM is instructed not to predict the speaker when it cannot be inferred from the input dialogue, it mostly selects a speaker from within the dialogue, resulting in higher recall for both all segments and segments from main characters.
LLR does not predict speakers for 6.8\% of the test set, while our method classifies only 0.77\% as \unknown.

\paragraph{Precision-POCS curve.}
We demonstrate the Precision -- Proportion of Classified Segments (POCS) trade-off in \cref{fig:precision_pocs}, showing results for all segments (left), long segments (middle), and short segments (right) by varying the threshold $D$. 
We include the curve from the LLR method for comparison.
The graphs show that the precision of character recognition decreases as we classify more audio segments. 
Compared to LLR, our method shows similar performance on long segments but demonstrates superior ability in classifying short segments. 
This verifies our method's effectiveness in identifying speakers of short utterances.

\subsection{Exemplar recognition accuracy}
\label{subsec:exemplar_recognition_acc}
\cref{tab:exemplar_acc} shows the accuracy of character recognition for the  exemplars.
We demonstrate that the accuracy of the audio exemplar building stage is nearly perfect. 
Out of 1,734 exemplars, most characters show 100\% accuracy. 
The pipeline mispredicts only \textbf{6} speakers (0.34\%), 5 from Frasier and one from Seinfeld. 
This high accuracy offers two advantages: (i) more audio segments are correctly classified, and (ii) more precise embeddings are obtained for local embedding classification of short segments. 
In terms of the number of exemplars, we extract more audio samples than \cite{Korbar24} from the same Seinfeld test set (609 vs. 407). 

\begin{table}[tb]
\centering
\caption{Exemplar recognition performance on LLR-TV. \textbf{\# exem} denotes the number exemplars extracted from our method and \textbf{\# correct} denotes the number of correctly classified exemplars. \textbf{Others} are a group of characters who are usually guest stars for one or a few episodes and the number of them is given in parentheses.}
\label{tab:exemplar_acc}
\resizebox{\linewidth}{!}{
\begin{tabular}{@{}lccclccclccc@{}}
\toprule
\multicolumn{4}{c}{\textbf{Frasier}}                                                           & \multicolumn{4}{c}{\textbf{Seinfeld}}                                                          & \multicolumn{4}{c}{\textbf{Scrubs}}                                                            \\ \midrule
\multicolumn{1}{c}{\textbf{Char}} & \textbf{\# exem} & \textbf{\# correct} & \textbf{Acc (\%)} & \multicolumn{1}{c}{\textbf{Char}} & \textbf{\# exem} & \textbf{\# correct} & \textbf{Acc (\%)} & \multicolumn{1}{c}{\textbf{Char}} & \textbf{\# exem} & \textbf{\# correct} & \textbf{Acc (\%)} \\ \midrule
\textbf{Frasier}                  & 347              & 342                 & 98.6              & \textbf{Jerry}                    & 353              & 352                 & 99.7              & \textbf{J.D.}                     & 152              & 152                 & 100               \\
\textbf{Niles}                    & 58               & 58                  & 100               & \textbf{George}                   & 41               & 41                  & 100               & \textbf{Dr.Cox}                   & 102              & 102                 & 100               \\
\textbf{Roz}                      & 28               & 28                  & 100               & \textbf{Elaine}                   & 66               & 66                  & 100               & \textbf{Turk}                     & 22               & 22                  & 100               \\
\textbf{Daphne}                   & 30               & 30                  & 100               & \textbf{Kramer}                   & 27               & 27                  & 100               & \textbf{Dr.Kelso}                 & 65               & 65                  & 100               \\
\textbf{Martin}                   & 31               & 31                  & 100               & \textbf{}                         &                  & \textbf{}           & \textbf{}         & \textbf{Elliot}                   & 88               & 88                  & 100               \\
\textbf{}                         &                  &                     &                   & \textbf{}                         &                  &                     &                   & \textbf{Carla}                    & 73               & 73                  & 100               \\ \midrule
\textbf{Others (7)}               & 47               & 47                  & 100               & \textbf{Others (12)}              & 122              & 122                 & 100               & \textbf{Others (13)}              & 82               & 82                  & 100               \\ \bottomrule
\end{tabular}}
\end{table}

\subsection{Qualitative examples}
\label{subsec:results_qual}
\begin{figure}[t]
  \centering
  \includegraphics[width=\linewidth]{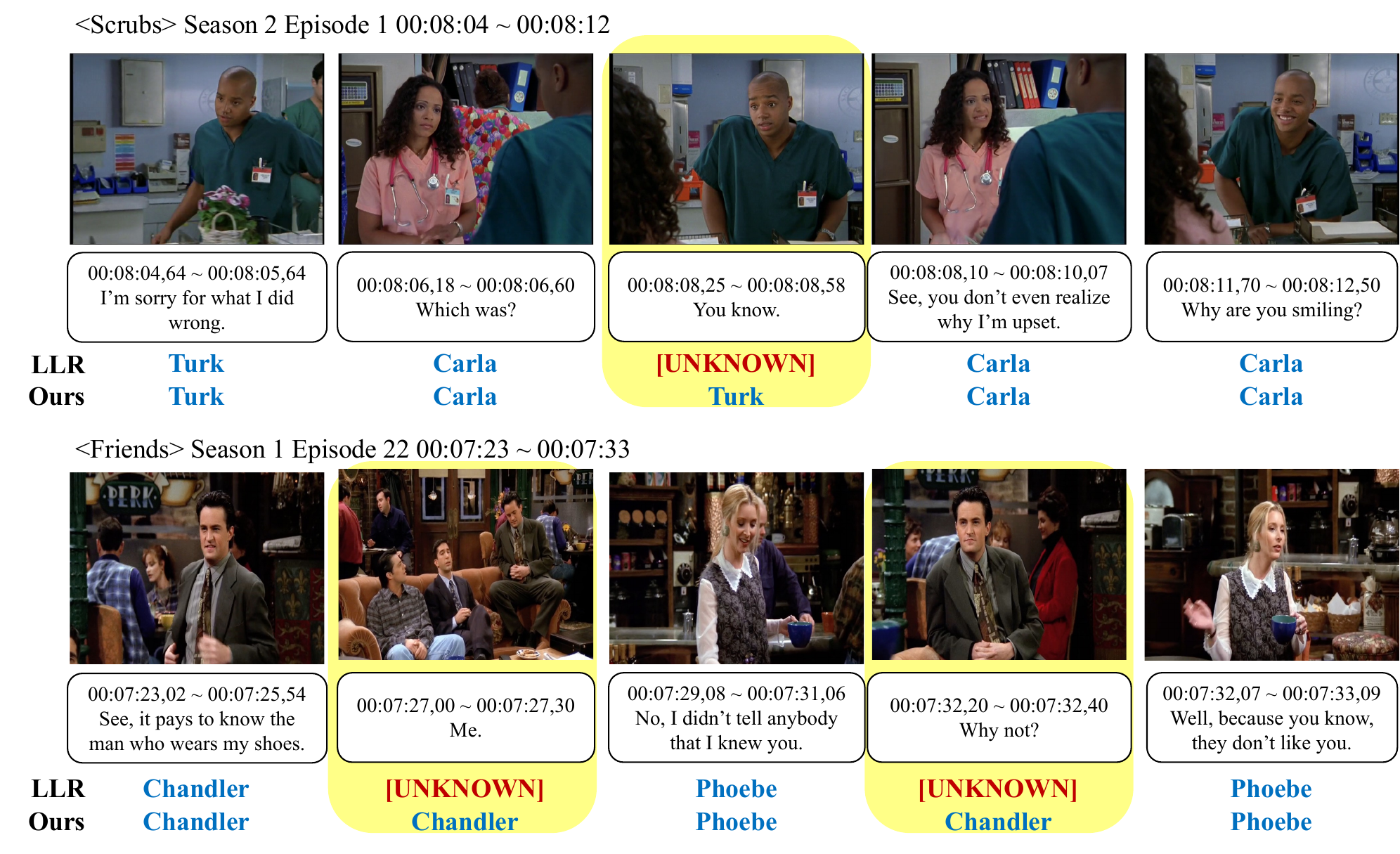}
  \caption{Qualitative examples from two TV series, \textit{Scrubs} and \textit{Friends}.}
  \label{fig:qual_examples}
\end{figure}

We present qualitative results from two series, \textit{Scrubs} and \textit{Friends}, in \cref{fig:qual_examples}. 
The figure shows speech recognition output and corresponding timestamps produced by our method, along with character recognition results from both LLR and this approach.
In both series, LLR fails to predict speakers for short utterances such as \textit{"You know."}, \textit{"Me."} or \textit{"Why not?"}.
In contrast, our method utilises the temporal context of the conversation to correctly classify the speaker for these brief segments.
It is important to note that the yellow utterances in the figure are initially classified as \unknown.
However, after employing LLM, these are correctly assigned to the appropriate speakers.

\section{Conclusions}
\label{sec:conclusions}

This paper presents an advanced framework for character-aware audio-visual subtitling in TV shows, addressing limitations in existing methods. 
Key contributions include a novel method for identifying speakers in short segments using temporal context, the use of local visual embeddings around lip-moving areas, and validation on a large dataset covering 12 TV series. 
Results demonstrate significant improvements in both diarisation performance and character recognition accuracy, particularly for short speech segments.

\begin{credits}
\subsubsection{\ackname}
This research is supported by EPSRC Programme Grant VisualAI EP/T028572/1 and a Royal Society Research Professorship RP\textbackslash R1\textbackslash 191132. We thank Robin Park and Bruno Korbar for helpful discussions.

\end{credits}
%
%
\bibliographystyle{splncs04}
\bibliography{shortstrings, vgg_local, vgg_other, ref}
\clearpage
\appendix
\title{Character-aware audio-visual subtitling in context -- supplementary material} 
\author{Jaesung Huh \and
Andrew Zisserman}
\institute{Visual Geometry Group, Department of Engineering Science, University of Oxford \\
\email{\{jaesung,az\}@robots.ox.ac.uk}}
\authorrunning{J.Huh and A.Zisserman}
\maketitle
\section{Character recognition accuracy on \bazingadataset}

\cref{tab:characc_bazinga} shows the character recognition performance on \bazingadataset, comparing with that from LLR~\cite{Korbar24}.
The reported performance of LLR corresponds to its highest character recognition accuracy (\textbf{Acc.}).
This peak accuracy is achieved by adjusting the threshold during the nearest centroid classification process.
We showcase that our method achieves higher accuracy than LLR in all TV shows.
It also achieves higher recall for both `all characters' and `main characters' compared to LLR.
LLR exhibits higher precision for 'all characters' and 'main characters' in a few TV shows.
This difference in performance is due to how each method handles short segments. 
LLR does not assign speakers to some of short segments, instead marking the character as \unknown. 
In contrast, our LLM tends to predict speakers for these \unknown\ segments after local embedding classification. 
It often assigns one of the characters appearing in the dialogue, even when we explicitly instruct the LLM that it doesn't have to make an assignment if it's unsure.
Thus, the precision decreases but the recall increases.

\begin{table}[!th]
\centering
\caption{Character recognition performance on \bazingadataset\ test set.  \textbf{Acc.} is the character recognition accuracy for those which overlap with one of the groundtruth timestamps. \textbf{Prec.} and \textbf{Rec.} indicate the precision and recall of overall audio segments respectively, while \textbf{Prec.(M)} and \textbf{Rec.(M)} are those of main characters in TV shows.}
\label{tab:characc_bazinga}
\resizebox{\linewidth}{!}{
\begin{tabular}{@{}lccccc|ccccc@{}}
\toprule
& \multicolumn{5}{c|}{LLR~\cite{Korbar24}}                                                    & \multicolumn{5}{c}{\textbf{Ours}}                                                                 \\ \midrule
& \textbf{Acc.} & \textbf{Prec.} & \textbf{Rec.} & \textbf{Prec.(M)} & \textbf{Rec.(M)} & \textbf{Acc.} & \textbf{Prec.} & \textbf{Rec.} & \textbf{Prec.(M)} & \textbf{Rec.(M)} \\ \midrule
\textbf{B.G.}    & 62.7         & 67.7           & 64.1          & \textbf{72.7}     & 71.1             & \textbf{68.7}     & \textbf{69.1}      & \textbf{70.1}     & 69.1              & \textbf{75.9}    \\
\textbf{B.B.}          & 60.5         & 69.4           & 61.4          & \textbf{83.6}     & 64.9             & \textbf{67.5}     & \textbf{72.6}      & \textbf{68.6}     & 76.3              & \textbf{69.4}    \\
\textbf{Buffy} & 57.6         & 62.4  & 58.1          & 71.7              & 57.5             & \textbf{61.6}     & \textbf{62.6}      & \textbf{62.1}     & \textbf{75.0}     & \textbf{65.0}    \\
\textbf{Friends}               & 70.9         & \textbf{82.1}  & 72.9          & \textbf{85.8}     & 79.7             & \textbf{74.2}     & 75.6               & \textbf{76.4}     & 76.0              & \textbf{83.2}    \\
\textbf{GoT}       & 52.6         & 55.4           & 55.7          & \textbf{57.3}     & 59.7             & \textbf{61.6}     & \textbf{62.9}      & \textbf{65.3}     & 54.1              & \textbf{63.7}    \\
\textbf{Lost}                  & 51.4         & 61.2           & 53.4          & 62.6              & 56.6             & \textbf{60.4}     & \textbf{63.9}      & \textbf{62.6}     & \textbf{65.4}     & \textbf{67.3}    \\
\textbf{TBBT}   & 80.2         & \textbf{88.8}  & 80.4          & \textbf{90.9}     & 85.4             & \textbf{81.5}     & 82.7               & \textbf{81.8}     & 83.2              & \textbf{86.2}    \\
\textbf{Office}            & 68.7         & \textbf{77.5}           & 68.9          & \textbf{84.8}     & 74.1             & \textbf{70.8}     & 74.4      & \textbf{71.2}     & 75.4              & \textbf{75.0}    \\
\textbf{W.D.}      & 49.7         & \textbf{60.3}  & 51.9          & \textbf{57.0}     & 52.9             & \textbf{54.7}     & 57.6               & \textbf{57.1}     & 53.5              & \textbf{64.0}    \\ \bottomrule
\end{tabular}}
\end{table}

\section{Effects of utilising spatial regions and speech enhancement on audio exemplar yield and performance}
\label{sec:results_cropse}

\cref{tab:effect_cropdenoise} demonstrates how cropping lip areas and speech enhancement improve exemplar yield. 
Unlike LLR, which uses whole frames without speech enhancement, our method identifies characters by focusing on the spatial region of speakers and reducing background noise to decrease false positive peaks, increasing exemplar yield by 8.5\% and 0.9\% respectively.
This showcases the effectiveness of our cropping-based approach in terms of exemplar yield.

\begin{table}[t]
\centering
\caption{Effect of using local visual predictions (vis) and speech enhancement (SE) on exemplar yield on the LLR-TV.}
\label{tab:effect_cropdenoise}
\resizebox{\linewidth}{!}{
\begin{tabular}{@{}lcccccc@{}}
\toprule
\multicolumn{1}{c}{}                & \multicolumn{2}{c}{LLR}       & \multicolumn{2}{c}{LLR + vis} & \multicolumn{2}{c}{LLR + vis + SE (Ours)} \\ \cmidrule(l){2-7} 
\multicolumn{1}{c}{}                & \# of exemplars & \% of total & \# of exemplars   & \% of total   & \# of exemplars              & \% of total              \\ \midrule
1-1. VAD + ASR                      & 5554            & 100         & 5554              & 100           & 5554                         & 100                      \\ \midrule
1-2. Audio-visual speaker recognition          & 2192            & 39.5        & 3332              & 60.0          & 3409                         & \textbf{61.4}                     \\ \midrule
1-3. Audio filtering                & 1213            & 21.8        & 1681              & 30.3          & 1734                         & \textbf{31.2}                     \\ \bottomrule
\end{tabular}}
\end{table}

\section{List of main characters per series}
\cref{tab:mainchar} lists the main characters per series, both in LLR-TV and \bazingadataset.
Note that we report the names of main characters in \bazingadataset\ as they are in the dataset annotation.

\begin{table}[h]
\caption{List of main characters per show.}
\label{tab:mainchar}
\resizebox{\linewidth}{!}{
\begin{tabular}{@{}ll@{}}
\toprule
\multicolumn{1}{c}{\textbf{Show}}         & \multicolumn{1}{c}{\textbf{Main characters}}                                                                                                                                                                \\ \midrule
\multicolumn{2}{l}{\textbf{LLR-TV}}                                                                                                                                                                                                                                \\ \midrule
\textbf{Frasier}                          & Frasier, Martin, Niles, Roz, Daphne                                                                                                                                                                         \\
\textbf{Seinfeld}                         & Jerry, Elaine, George, Kramer                                                                                                                                                                               \\
\textbf{Scrubs}                           & J.D., Dr.Cox, Dr.Kelso, Carla, Turk, Elliot                                                                                                                                                                 \\ \midrule
\multicolumn{2}{l}{\textbf{\bazingadataset}}                                                                                                                                                                                                                             \\ \midrule
\textbf{Battlestar Galactica (B.G.)}      & \begin{tabular}[c]{@{}l@{}}Admiral William Adama, President Laura Roslin, Captain Kara Thrace, Cpt. Lee Adama, Number six, \\ Dr. Gaius Baltar, Lt. Sharon Valerii\end{tabular}                             \\
\textbf{Breaking Bad (B.B.)}              & Walter White, Skyler White, Jesse Pinkman, Hank Schrader, Marie Schrader                                                                                                                                    \\
\textbf{Buffy the Vampire Slayer (Buffy)} & Buffy Summers, Willow Rosenberg, Xander Harris, Angel, Rubert Giles                                                                                                                                         \\
\textbf{Friends}                          & Rachel Green, Monica Geller, Phoebe Buffay, Joey Tribbiani, Chandler Bing, Dr. Ross Geller                                                                                                                  \\
\textbf{Game of Thrones (GoT)}            & \begin{tabular}[c]{@{}l@{}}Danerys Targaryen, Jon Snow, Jorah Mormont, Tyrion Lannister, Catelyn Stark, Sansa Stark, \\ Arya Stark, Cersei Lannister, Eddard Stark, Robert Baratheon\end{tabular}           \\
\textbf{Lost}                             & \begin{tabular}[c]{@{}l@{}}Dr. Jack Sheperd, Kate Austen, Sayid Jarrah, Hugo Reyes, Sunhwa Kwon, Charlie Pace, \\ Clarie Littleton, Michael Dawson, John Locke, Shannon Rutherford, James Ford\end{tabular} \\
\textbf{The Big Bang Theory (TBBT)}       & Sheldon Cooper, Penny, Howard Wolowitz, Raj Koothrappali, Leonard Hofstadter                                                                                                                                \\
\textbf{The Office (Office)}              & \begin{tabular}[c]{@{}l@{}}Jim Halpert, Michael Scott, Ryan Howard, Pam Beesly, Dwight Schrute, Stanley Hudson, \\ Phyllis Vance, Angela Martin\end{tabular}                                                \\
\textbf{Lost}                             & \begin{tabular}[c]{@{}l@{}}Rick Grimes, Lori Grimes, Carl Grimes, Carol Peletier, Shane Walsh, Andrea Harrison, \\ Dale Horvath, Glenn Rhee\end{tabular}                                                    \\ \bottomrule
\end{tabular}}
\end{table}

\section{LLM prompt}
\begin{table}[tb]
    \centering
    \caption{
    LLM prompt to determine \unknown.
    }
    \label{tab:prompt}
    \begin{tabular}{p{\linewidth}}
        \toprule
        \underline{\textbf{\textsc{Prompt for determining [UNKNOWN]}}} \\
        \textbf{system} : You are a AI assistant to analyze the transcript of TV shows. Your job is to figure out who are [UNKNOWN]s in a dialogue in TV shows. Tell the truth and answer as precisely as possible. \\

        \textit{(provide the dialogue here)}

        \textbf{user} : Write a summary for the above conversation.
        
        \textbf{assistant} : \textit{(model generates the summary)}
        
        \textbf{user} : Based on the summary, your job is to identify the name of the speaker of `[UNKNOWN]' when the line starts and ends with `**'. You must use the context and the flow of the dialogue, using the speakers' names and what they speak. The list of speakers with their corresponding tokens are provided below. Choose [UNKNOWN] if his or her name is not in the dialogue, or when you are not sure. \\
        \textit{(provide the list of speakers here)} \\
        Only output one number after ANSWER: \\

        \textbf{assistant} : ANSWER: \\ \\

        \underline{\textbf{\textsc{Example of the dialogue ($n_{LLM}=5$)}}} \\
                Dr.Cox : You can use it. \\
                Dr.Cox : God, I hate Halloween. \\
                Carla : Somebody needs to adjust their attitude if they want some candy. \\
                Dr.Cox : You mean, the popcorn balls and the deformed lollipops. \\
                Dr.Cox : I mean, honestly, where do you get this crap anyway? \\
                **[UNKNOWN] : I made it.** \\
                NurseRoberts : If you want name brand candy, my fish is packed with peanuts. \\
                Dr.Cox : Of course it is.\\
                Carla : Oh, what's the matter?\\
                Carla : Did Raggedy Ann scare you?\\
                Dr.Cox : What are you, a rat?\\ \\

        \underline{\textbf{\textsc{Example of the list of speakers}}} \\
        1: Dr.Cox, 2: Carla, 3: NurseRoberts, 4: [UNKNOWN] \\
        \bottomrule
    \end{tabular}
\end{table}

\cref{tab:prompt} shows the LLM prompt we've used to determine the \unknown\ character in the dialogue.
We adopt the strategy introduced in \cite{park2024automated}.
Given the query dialogue with the \unknown\ character we want to classify, we first ask the LLM to summarize the dialogue.
Then, we ask the model who would be the \unknown\ in the dialogue based on the generated summary.
A list of characters with their corresponding indices is provided in the prompt. We compute the softmaxed logits of the tokens corresponding to each index and choose the largest one to assign the speaker.
We use this same prompt for experiments in LLR-TV and \bazingadataset.

There can be multiple \unknown s in the query dialogue. 
We mark the sentence for which we want to identify the speaker by placing asterisks (**) before and after it. This explicitly instructs the LLM to predict the speaker only for that specific sentence.

\end{document}